\documentclass[sigconf]{aamas} 

\usepackage{comment}
\usepackage{hyperref}
\usepackage{listings}
\usepackage{nccmath}

\definecolor{keycolor}{RGB}{191,0,64}
\definecolor{neonpurple}{RGB}{128, 0, 255}

\lstdefinestyle{RDFStyle}{
basicstyle=\fontsize{7}{7}\selectfont\ttfamily\color{blue},
morekeywords=[2]{PREFIX,@prefix,eve:},
keywordstyle=[2]\color{keycolor},
moredelim=[s][\color{red}]{"}{"},
moredelim=[is][\color{keycolor}]{|>}{<|},
moredelim=[s][\color{neonpurple}]{<}{>},
numbersep=5pt,   
numberstyle=\color{black},
literate=*{[}{{\textcolor{black}{[}}}{1}
         {]}{{\textcolor{black}{]}}}{1}
         {;}{{\textcolor{black}{;}}}{1}
         {.}{{\textcolor{black}{.}}}{1}
}
\makeatletter
\let\orig@lstnumber=\thelstnumber
\newcommand\lstsetnumber[1]{\gdef\thelstnumber{#1}}
\newcommand\lstresetnumber{\global\let\thelstnumber=\orig@lstnumber}
\makeatother

\usepackage{balance} 

\renewcommand\footnotetextcopyrightpermission[1]{} 
\setcopyright{ifaamas}
\acmConference[AAMAS '23]{Proc.\@ of the 22nd International Conference
on Autonomous Agents and Multiagent Systems (AAMAS 2023)}{May 29 -- June 2, 2023}
{London, United Kingdom}{A.~Ricci, W.~Yeoh, N.~Agmon, B.~An (eds.)}
\copyrightyear{2023}
\acmYear{2023}
\acmDOI{}
\acmPrice{}
\acmISBN{}

\acmSubmissionID{1017}

\title[Signifiers as a First-class Abstraction in Hypermedia Multi-Agent Systems]{Signifiers as a First-class Abstraction\\in Hypermedia Multi-Agent Systems}

\author{Danai Vachtsevanou}
\affiliation{
  \institution{University of St. Gallen}
  \city{St. Gallen}
  \country{Switzerland}}
\email{danai.vachtsevanou@unisg.ch}

\author{Andrei Ciortea}
\affiliation{
  \institution{University of St. Gallen}
  \city{St. Gallen}
  \country{Switzerland}}
\email{andrei.ciortea@unisg.ch}

\author{Simon Mayer}
\affiliation{
  \institution{University of St. Gallen}
  \city{St. Gallen}
  \country{Switzerland}}
\email{simon.mayer@unisg.ch}

\author{Jérémy Lemée}
\affiliation{
  \institution{University of St. Gallen}
  \city{St. Gallen}
  \country{Switzerland}}
\email{jeremy.lemee@unisg.ch}

\begin{abstract}

Hypermedia APIs enable the design of reusable hypermedia clients that discover and exploit affordances on the Web. However, the reusability of such clients remains limited since they cannot plan and reason about interaction. 
This paper provides a conceptual bridge between hypermedia-driven affordance exploitation on the Web and methods for representing and reasoning about actions that have been extensively explored for Multi-Agent Systems (MAS) and, more broadly, Artificial Intelligence. 
We build on concepts and methods from Affordance Theory and Human-Computer Interaction that support interaction efficiency in open and evolvable environments to introduce \emph{signifiers} as a first-class abstraction in Web-based MAS: Signifiers are designed with respect to the agent-environment context of their usage and enable agents with heterogeneous abilities to act and to reason about action. We define a formal model for the contextual exposure of signifiers in hypermedia environments that aims to drive affordance exploitation. 
We demonstrate our approach with a prototypical Web-based MAS where two agents with different reasoning abilities proactively discover how to interact with their environment by perceiving only the signifiers that fit their abilities. 
We show that signifier exposure can be inherently managed based on the dynamic agent-environment context towards facilitating effective and efficient interactions on the Web.
\end{abstract}


\newcommand{\BibTeX}{\rm B\kern-.05em{\sc i\kern-.025em b}\kern-.08em\TeX}

\begin{document}


\pagestyle{fancy}
\fancyhead{}


\maketitle 


\section{INTRODUCTION}

Research on autonomous agents and Multi-Agent Systems (MAS) has been focusing on models and methods for the flexible exploitation of environmental resources. To engineer large-scale and afford- ance-rich MAS, the Web has been investigated as an enabler for autonomous agents to make the most of their reasoning and decision-making abilities in discovering and exploiting affordances of Web resources for their complex goals~\cite{ciortea2019decade}. The prime mechanism for affordance discovery and exploitation on the Web is \textit{hypermedia-driven interaction}, which drives interaction based on representations of combined high-level \textit{semantic information} and lower-level \textit{controls}~\cite{fielding_2008}: Agents would \emph{reason} about their actions based on semantic information, e.g., that grasping an item is possible because the item is close to a robotic arm, and \emph{act} by using the accompanying controls, e.g,. with an HTTP request to the arm's Web API. After acting, new hypermedia become available based on the ``Hypermedia as the Engine of Application State" (HATEOAS) principle of the Web architecture~\cite{fielding2000architectural}, enabling agents to reason about and to exploit new affordances, e.g., to pick the grasped item. 

The W3C Web of Things~\cite{charpenay2016introducing} already extends the Web environment with machine-readable specifications of hypermedia-driven APIs, namely \emph{Thing Descriptions}, which enable programming applications on top of domain-specific abstractions rather than low-level controls (e.g., for a specific application-layer protocol like HTTP). As a result, such applications become reusable hypermedia clients that exploit affordances offered by different physical and virtual entities on the Web. Yet, even such clients behave more like utility applications rather than autonomous users of hypermedia as their lack of adaptivity limits their reusability: Clients search for specific semantic information whose discovery triggers the exploitation of controls based on some pre-compiled logic, comparable to an agent that -- as a reflex -- always picks items that are close to it.

This compares with human users who \textit{autonomously} and flexibly \textit{take the initiative} to exploit new affordances and dynamically achieve their goals, by \textit{exploring the environment} and  \textit{interpreting, reasoning about, and planning on top of signifiers}: cues that represent high-level semantic information about affordances and that are specifically designed to increase the interpretability and discoverability of affordances based on the principles of Human-Computer Interaction (HCI)~\cite{lemee2022sign}, and exposed in a way that complements the run-time human-environment context. As a result, human users effectively and efficiently engage in flexible affordance discovery and exploitation even within new and affordance-rich environments. 

Hence, compared to humans, hypermedia clients today lack a) awareness of entities like signifiers, which could tactically drive their interactions in the Web environment (e.g. through HATEOAS), and b) the abilities to reason, plan, and proactively deliberate about their actions based on such perceived environmental entities. 

In this paper, we introduce signifiers as a first-class abstraction in hypermedia-based MAS as a means of provisioning machine-readable interaction specifications for clients that reason and act as autonomous agents on the Web. Inspired by the interaction principles of HCI, we aim at a user-centered design of signifiers that directly accommodate the versatility of hypermedia environments and of autonomous agents and their abilities and objectives. Concretely, we define a formal model of a mechanism for the contextual exposure of signifiers in the hypermedia environment to support interaction effectiveness and efficiency based on HATEOAS.

To demonstrate our approach, we implemented a prototypical  hypermedia-based MAS, where two agents with different types of reasoning abilities proactively discover how to interact with shared artifacts in the environment, by dynamically perceiving  signifiers in a manner that suits their abilities and context. We show that the exposure of such interaction specifications to autonomous agents can be inherently managed with respect to the dynamic agent-environment context towards facilitating effective and efficient interactions on the Web.


\section{BACKGROUND AND RELATED WORK}

We provide an overview of related work about how affordance exploitation by \textit{human agents} (Sect.~\ref{subsec:hci}) compares to affordance exploitation by \textit{hypermedia clients} (Sect.~\ref{subsec:clients}), and how research on \textit{MAS} could be used to reduce the gap between the two (Sect.~\ref{subsec:agents}).

\subsection{Human-centered Interaction Design} \label{subsec:hci}

Affordance Theory~\cite{gibson1977theory} examines how animals control their behavior by perceiving and exploiting \textit{affordances} in their environment. 
Chemero defines an \textit{affordance} as a behavior possibility that is a relationship between a) an \textit{ability} of an agent and b) a \textit{situation} that includes agents and features of the environment~\cite{chemero2007complexity}. The ability and the situation of an affordance are considered \textit{complementary} to each other, i.e., the presence of an agent with an ability within a situation makes an affordance \textit{exploitable} by the agent. For example, the affordance \texttt{graspable} of an item becomes exploitable by a human agent if the agent has the \texttt{ability-to-reach} the item in the \texttt{hand-is-empty} situation. An affordance $aff$ is formally defined in~\cite{chemero2007complexity} as:
\begin{equation}
\label{eq:aff}
    aff = <a, s>,
\end{equation}
where $a$ is an agent  ability and $s$ is an agent-environment situation whose simultaneous presence make $aff$ exploitable by an agent.

When an affordance becomes exploitable by an agent, the agent has the \textit{ability} to perform a \textit{behavior}. For example if an item is indeed \texttt{graspable}, then the human agent  can exhibit the \texttt{ability-to- grasp} the item, i.e. to perform the behavior \texttt{grasp}. Therefore, an ability $a'$ is defined as the quality of an agent to perform a behavior $b$ when an affordance $aff$ is exploitable;  formally defined as:
\begin{equation}
\label{eq:ability}
    a' = <aff, b>,
\end{equation}
where $aff$ is an affordance (cf. Def.~\ref{eq:aff}) and $b$ is the behavior that the agent has the ability to perform when $aff$ is exploitable.

Ability $a'$ may in turn be complementary to another situation $s'$, such that the simultaneous presence of $a'$ and $s'$ makes another affordance $aff'$ exploitable by the agent, i.e. $aff' = <a', s'>$. For example, the \texttt{ability-to-grasp} can be considered as complementary to the situation \texttt{nothing-on-top-of-the--item}. Such complementarity can be defined through the affordance $aff'$ that is the affordance \texttt{pickable} of the item.

Through Defs.~\ref{eq:aff} and \ref{eq:ability}, affordances and abilities are impredicatively defined in terms of one another, forming a complex system that can be studied through the use of hypersets~\cite{chemero2007complexity}. For this, a system is modeled through sets of affordances and abilities to examine how (human) agents can control their behaviors in the system: Consider the set $B$ of the behaviors that can be performed in the system, $B=\{b_1, b_2,..., b_n\}$ (e.g, $B$ = $\{grasp, reach, open\mbox{-}door, ...\}$). Then, the affordances and abilities that are considered for $B$ are, respectively, included in the sets $Affords\mbox{-}B$ and $Ability\mbox{-}to\mbox{-}B$. These sets are impredicatively inter-defined as follows:
\begin{equation}
\label{eq:affords-b}
     Affords\mbox{-}B = \{<a_1, s_1> <a_2, s_2>, ..., <a_n, s_n>\},
\end{equation}
\begin{equation}
\label{eq:abilityto-b}
     Ability\mbox{-}to\mbox{-}B = \{<aff_1, b_1>, <aff_2, b_2>, ..., <aff_n, b_n>\}\footnote{For example, $Ability\mbox{-}B= \{<graspable, grasp>, <reachable, reach>, ...\}$, and $Affords\mbox{-}B= \{<ability\mbox{-}to\mbox{-}reach, item\mbox{-}is\mbox{-}reachable>, ...\}$.}.
\end{equation}

Affordances have provided valuable input to the domain of Human-Computer Interaction (HCI) towards the design of human-made things that increase the \textit{discoverability} and \textit{interpretability} of their offered affordances. Specifically, Norman introduced \textit{signifiers} to discuss the design of \textit{perceivable cues that can be interpreted meaningfully to reveal information about affordances}~\cite{norman2008way,norman2013design}. Specifically, Norman advocates the \textit{human-centered} design of signifiers that should be driven by the needs, objectives, and abilities of targeted users. Then, signifiers become environmental cues that can be \textit{intuitively and reliably} discovered and interpreted to reveal \textit{what} are the possible behaviors and \textit{how} these behaviors can be performed. 

The methodical design \textit{and exposure} of signifiers in the environment help human agents to explore, reason about, and exploit affordances even when situated within large-scale and affordance-rich settings, such as in the hypermedia environment of the Web. On the Web, signifiers are \textit{designed with respect to their expected human users} -- abstracting away from low-level interaction details that do not concern the users (i.e. the hypermedia controls), and revealing only the information that is required to intuitively invite user behavior (i.e., the informational part of hypermedia)\footnote{Although Affordance Theory defines affordances in relation to their functional aspects, in HCI functional affordances are commonly differentiated from cognitive ones~\cite{hartson2003cognitive} to separate design concerns, for instance between a service provider and a UI designer.}. On top of this, the \textit{exposure of signifiers is continuously adjusted} to better serve the dynamic agent-environment context of affordance exploitation. On the Web, this is achieved through the HATEOAS principle, which can be viewed as the environmental (server-side) support that manages to keep the provision of affordances and signifiers in alignment with the current application state (and the user needs that are expected in the current application state).

\subsection{Engineering Hypermedia Clients}\label{subsec:clients}

Hypermedia clients are applications capable of hypermedia-driven interaction, i.e. they exploit affordances and advance the application state by using hypermedia controls, such as \textit{links} and \textit{forms}, while application logic is coupled to the information that accompanies the controls, such as \textit{identifiers} of link relations or of form operation and content types. Then, through HATEOAS, each interaction results in a client receiving new information in the form of hypermedia regarding the (newly) available affordances. As a result, hypermedia clients remain decoupled from specific Web resources and data objects, and are thus more resilient to API changes than clients which are specialized and tailored to specific Web APIs~\cite{amundsen2017restful}. 

In case decisions about action are made by a human agent, hypermedia clients become as generic (and reusable) as Web browsers are today: Clients iteratively 1) parse and present signifiers to human agents (see Sect.~\ref{subsec:hci}), 2) wait for human agents to reason and act based on exposed signifiers towards achieving their objectives, and 3) use hypermedia controls to advance the application state based on the human agent's decisions. In case the client itself has some higher-level objective to achieve, signifiers need to carry unambiguous semantics so that they can be handled reliably by the machine. This can be achieved through \textit{standardized identifiers} of link relations and operation types. For example, the Hydra Core Vocabulary~\cite{lanthaler2013hydra} and the Hypermedia Controls Ontology (HCTL)\footnote{The HCTL Ontology is available online: \url{https://www.w3.org/2019/wot/hypermedia}} enable the semantic description of RESTful Web APIs, thus offering a foundation for examining how to model signifiers for reusable clients with more complex application logic; subsequently, the W3C Web of Things (WoT) ~\cite{Kovatsch:20:WTT} examines how to enable clients to cope with the increasing number of heterogeneous devices that are being connected to the Internet. For this, W3C WoT takes a step further from hypermedia controls modelling (reusing HCTL) by providing an easy-to-use interaction model for defining \emph{Thing Descriptions (TDs)}, i.e. machine-readable specifications that reveal higher-level interaction semantics of device (\textit{Thing}) affordances.

Although WoT TDs enable the reuse of client application logic, there is also research on interaction specifications that allow for more adaptive hypermedia clients. For example, clients in~\cite{kafer2018rule} are executable Linked Data specifications that contain N3 rules\footnote{The N3 specification is available online: \url{https://www.w3.org/DesignIssues/Notation3}} for the more contextual consumption of hypermedia (e.g. currently exploitable affordances can be identified), and can be integrated in Hierarchical Task Network workflow specifications to manifest more complex behaviors~\cite{assfalg2021integrated}. Finally, interaction specifications in the RESTdesc format~\cite{verborgh2012functional} can be used as input to N3 reasoners, thus enabling the more dynamic combination of affordances based on the run-time system objectives. 

\subsection{Reasoning for Goal-driven Interaction} \label{subsec:agents}
\textit{Autonomous agents} are capable of adaptive behavior, since they handle abstractions that enable them to reason about their actions and take the initiative to interact with respect to their dynamic goals. For example, agents that implement the Procedural Reasoning System~\cite{georgeff1987reactive} can discover and execute behaviors in the form of \textit{plans} and reason about the applicability and relevance of such plans based on their own \textit{beliefs}, \textit{desires}, and \textit{intentions}~\cite{rao1995bdi}. Additionally, autonomous agents can synthesize new plans using methods of automated planning~\cite{ghallab2016automated}, by reasoning on the \textit{preconditions} and \textit{postconditions} of actions with respect to desired effects. 

Specifically, behaviors that relate to affordance exploitation are possible because the above abstractions establish a relationship between the agents and their environment. For example, the Agents \& Artifacts meta-model (A\&A)~\cite{ricci2011environment} defines \textit{artifacts} as tools organized within \textit{workspaces}, through which agents can perceive and manipulate the environment. Then, interaction \textit{efficiency} can be achieved through the agents' ability to be \textit{situated} within (logical) workspaces, and hence to direct the \textit{scope} of their perception and action towards co-located artifacts of interest.

Autonomous agents and their ability to reason about action with respect to  goals and environmental artifacts have been also examined in the context of the Web, so as to enable the deployment of large-scale and affordance-rich Web-based Multi-Agent Systems (MAS). Hypermedia MAS~\cite{ciortea2018engineering} are a class of MAS that remains aligned with the architectural principles of the Web towards endowing hypermedia clients with the abilities to cope with evolvable Web APIs. Agents in Hypermedia MAS can already discover and exploit affordances in dynamic hypermedia environments through WoT TDs~\cite{ciortea2018engineering}, but they are still not able to make the most of their reasoning and decision-making abilities since TDs are not specialized to establish a relationship to abstractions in agent-oriented programming and MAS (e.g. goals, action preconditions, etc.). 

\section{Signifiers for Hypermedia MAS}
In the following, we introduce signifiers as a first-class abstraction in Hypermedia MAS and provide a formalization of signifiers for this purpose in Sect.~\ref{subsec:signifiers}. The primary responsibility of signifiers is to support the discoverability and interpretability of affordances. Hence, in Sect.~\ref{subsec:sem}, we introduce a mechanism that exploits the agent-environment complementarity to expose only those signifiers that are relevant to agents situated in a hypermedia environment. Finally, in Sect.~\ref{subsec:srm}, we discuss how signifiers can be customised to support agents with different abilities, and we present signifiers for agents based on two well-known methods for reasoning about action -- the Procedural Reasoning System~\cite{georgeff1987reactive} and STRIPS~\cite{fikes1971strips}. 

\subsection{Agent-centered Design of Signifiers for Hypermedia-driven Interaction}\label{subsec:signifiers}
In this section, we define a general model for signifiers that a) capture the agent-environment complementarity required to exploit affordances, and b) enable the hypermedia-driven exploitation of affordances on the Web. 
Specifically, this model supports the interactions of autonomous agents in Hypermedia MAS.

\subsubsection{A General Model for Signifiers}
Our model for signifiers builds on top of the affordance model presented by Chemero and Turvey in ~\cite{chemero2007complexity} (see Sect.~\ref{subsec:signifiers}). We, accordingly, identify:
\begin{itemize}
    \item an affordance $aff_1=<a_1, s_1>$, e.g. the affordance \texttt{gripper- closable} of a robotic arm, which is exploitable when an autonomous agent has the ability $a_1$ to \emph{log in as an operator} by using the device's HTTP API in the situation $s_1$ that the \emph{gripper is open} (cf. Def.~\ref{eq:aff});
    \item an ability $a_0=<aff_1, b_1>$, e.g. the agent's ability to perform the behavior $b_1$ of \emph{closing the gripper} by using the device's HTTP API, when the affordance $aff_1$ \texttt{gripper-closable} is exploitable (cf. Def.~\ref{eq:ability});
    \item the related sets of affordances $Affords\mbox{-}B$ (cf. Def.~\ref{eq:affords-b}) and abilities $Ability\mbox{-}to\mbox{-}B$ (cf. Def.~\ref{eq:abilityto-b}) for examining how an agent can exploit affordances in the system based on the set of behaviors $B=\{b_1, b_2, ..., b_n\}$, e.g., $B= \{gripper\-\mbox{-}closable, login, ...\}$.
\end{itemize}

We define a \textit{signifier}  $sig$ as a perceivable cue or sign that can be interpreted meaningfully to reveal information about an affordance $aff_1$, formally defined as follows:
\begin{equation} \label{eq:signifier}
sig = (sp_b, A, c, salience),
\end{equation}
where:
\begin{itemize}
    \item $sp_b$ is the signified \emph{specification} of a behavior $b$ of the set $B$, i.e. the specification of a course of action that can be performed by an agent when $aff_1$ is exploitable;
    \item $A$ is a \emph{set of abilities}, where an ability is a quality of an agent to perform a behavior of the set $B$. Hence, $sig$ \emph{recommends} that agents should have all the abilities of the set $A$ so that $aff_1$ becomes exploitable. 
    \item $c$ is the \emph{context} which $sig$ recommends to hold so that $aff_1$ becomes exploitable, i.e. constraints to which the agent- environment situation is recommended to conform;
    \item $salience$ is the quality of $sig$ that indicates how useful or relevant $sig$ is.
\end{itemize}

A machine-understandable signifier in RDF~\cite{Lanthaler:14:RCA} is structured based on Def. \ref{eq:signifier} in Lst. \ref{lst:sg-basic} (l.7-13): It signifies the specification of the behavior \texttt{close-gripper}, recommended to be performed when an agent is able to become an operator and the gripper is open.

Note that the definition (and the representation) of a signifier (here, $sig$) does not include any direct reference to an affordance (here, $aff_1$). This is preferable for two reasons: First, affordances emerge upon the presence of \textit{individual agents} in the appropriate situation, while signifiers are defined with respect to \textit{agent types}. This is essential because designers typically do not have prior knowledge of the individual agents that will inhabit the environment. In this way, Def. \ref{eq:signifier} preserves the evolvability and reusability of signifiers, since agents \emph{remain more loosely coupled} to their environment. Second, \emph{affordances} are defined in Affordance Theory for studying how \textit{animals} control their behaviors. However, we model \emph{signifiers} for supporting interactions of \textit{autonomous agents}, which exhibit more heterogeneous cognitive and sensorimotor abilities than animals -- including agents whose cognitive abilities heavily rely on computational processes such as reasoning over representations, and whose sensorimotor abilities\footnote{In Affordance Theory, action is considered to be strictly coupled to perception. Although there are examples of autonomous agents which exhibit similar cognitive and sensorimotor abilities for performing perception-to-action behaviors, such as Rodney Brooks' physically embodied agents that implement the subsumption architecture~\cite{brooks1999cambrian}, a Hypermedia MAS is expected to feature a greater diversity of agents' abilities.} allow for behaviors that are much broader than animal behaviors. 

Even though there is no formal relation between signifiers and affordances, the designer of a signifier $sig$ may use the elements of Def. \ref{eq:signifier} to establish a (direct or indirect) relationship to specific elements of $aff_1$. The forms that this relationship can take can be narrowed down to the following cases:
\begin{itemize}
    \item[(C1)] $sp_b$ of the behavior $b$ is equivalent to the specification of the behavior $b_1$ for exploiting $aff_1$ and optionally any behaviors that lead to making $aff_1$ exploitable, i.e. $sp_b$ specifies behaviors of the set $B'\subseteq B$, and $B'	\supseteq \{b_1\}$;
    \item[(C2)]$A$ is equivalent to the set of any of the abilities that lead to making $aff_1$ exploitable, i.e. $A \subseteq Ability\mbox{-}to\mbox{-}B$;
    \item[(C3)] $c$ is equivalent to (part of) the constraints that are satisfied in situation $s_1$.
\end{itemize} 
C1-C3 should be used based on the experience and expectations that environment designers hold about targeted agents towards enabling effective and efficient interactions. Therefore, designers are responsible for choosing when to aggregate or omit to refer to information about the sets $Affords\mbox{-}B$ and $Ability\mbox{-}to\mbox{-}B$. 
For example, (C1) allows for managing the granularity of behavior $b$ that is specified in the signified specification $sp_b$. Considering that $aff_1$ is the affordance \texttt{gripper-closable}, then $b$ could be specified as the sequence $<b_2,b_1> = <login, close\mbox{-}gripper>$. Such granularity could be desirable if $b_2$ and $b_1$ are expected to be frequently performed in sequence or for enabling agents to access specifications of higher-level behaviors. Similarly, (C2) enables designers to avoid including in $A$ these abilities of $Ability\mbox{-}to\mbox{-}B$ that targeted agents are expected to always have. (C3) enables recommending a context $c$ by determining which constraints address the most significant aspects of situation $s_1$ that makes $aff_1$ exploitable. Generally, the context may capture constraints about agents and constraints about the entity offering an affordance, which are disjoint (e.g., constraints on the intentions of the agent, and respectively the state of the artifact) or interdependent (e.g., that the agent is situated in the same working environment that contains the environmental).

In our proposed model of signifiers, recommended abilities and context are not meant to \emph{regiment} agents' behaviors, i.e. to constrain whether an agent will actually exploit an affordance. However, they can both be beneficial in evaluating the relevance of an affordance exploitation, and consequently the relevance of the signifier for an agent with given abilities in a given situation. 

\label{lst:sg-basic}
\begin{lstlisting}[float=b, style=RDFStyle, caption=A (generic) signifier that reveals information about the affordance gripper-closable of a robotic arm., label=lst:sg-basic, numbers=left, xleftmargin=14pt, mathescape=true ]
@base <http://ex.org/wksp/1/arts/1>.
@prefix hctl:<https://www.w3.org/2019/wot/hypermedia#>.
$\lstsetnumber{\ldots}$... $\lstresetnumber\setcounter{lstnumber}{6}$
<#sig> a hmas:Signifier ;
  hint:signifies <#close-gripper> ;
  hint:recommendsAbility [ a manu:OperatorAbility ] ;
  hint:recommendsContext <#env-context> .
  
<#env-context> a hint:Context; sh:targetNode ex:leubot ;
  sh:property [ sh:path manu:hasGripperValue  ;
    sh:hasValue "500"^^xsd:integer ] .
  
<#close-gripper> a hint:ActionSpecification;
 hint:hasForm [ hctl:hasTarget leubot:base ;
  hctl:forContentType "application/json" ] ;
 hint:expects [ a hint:Input; 
  hint:hasSchema <#gripper-schema> ] .

<#gripper-schema> a js:ObjectSchema ;
 js:properties [ a js:IntegerSchema ;
  js:propertyName "manu:hasGripperValue"; js:enum "0"]] .
\end{lstlisting}

We further develop our model to define how \textit{behaviors} can be specified for \textit{hypermedia-driven interactions} in Hypermedia MAS, e.g., in the form of an AgentSpeak plan~\cite{bordini2007programming}, a JADE behavior~\cite{bellifemine2007developing}, etc. Due to this diversity, the only requirement that we formally impose for the definition of a behavior specification is that it specifies at least one action.

 An \textit{action} is specified through a) \textit{forms} that describe the hypermedia controls that can be used to implement and execute the action and b) optionally an \textit{input}. For example, the behavior \texttt{close- gripper} could be specified as an action through a form that describes an HTTP request, and a gripper input value. An \textit{action execution} can be treated as a behavior that is expected to result in a modification of the state of the environment (e.g., of the gripper). In this case, any entities in the system that monitor an agent executing an action, may perceive that the agent acts on the environment. However, an action could also be used for perception in the case in which an agent executes the action with the purpose of affecting its perception.
Then, a specification $sp_b$ formally specifies such a behavior $b$ (an action execution) as follows\footnote{We insert $\lfloor$ and $\rfloor$ to delimit expressions considered optional.}:
\begin{equation}
    sp_b = (Forms, \lfloor Input \rfloor), 
\end{equation}
where $Forms$ is a set of forms where each form describes an implementation of the specified action, and $Input$ is the input expected when it is possible or required to parameterize the action execution. List. \ref{lst:sg-basic} (l.15-24) captures a signified specification of \texttt{close-gripper}.

\subsubsection{Abilities of Autonomous Agents} 
In the following, we discuss  
how \textit{abilities} recommended by signifiers can facilitate the effective and efficient affordance discovery and exploitation in the MAS. For example, an agent reasoning on signifiers could infer that it should exploit a related affordance only if it has the appropriate abilities. Even if the agent does not currently exhibit the recommended abilities, it could still use this information towards extending its abilities, or delegating goals to capable agents. For example, if an agent $ag_1$ has the abilities of the set $A_1$ and is aware of a signifier $sig_1=(sp_b, A, c, salience)$, where $A \nsubseteq A_1$, then $ag_1$ may decide to forward $sig_1$ and request the performance of the related behavior from an agent $ag_2$ that has the abilities of the set $A_2$, where $A \subseteq A_2$. Next, we consider the abilities of agents to illustrate diverse aspects of agents that signifier designers could take into account.

An agent may have an \textit{ability to reason on a representation of its internal state}; e.g., a BDI agent can reason about its beliefs, desires, and intentions~\cite{rao1995bdi}. Recommending such abilities abilities ensures that agents have the appropriate \emph{cognitive skills} to perform a specified behavior. For instance, a signifier that signifies the specification of sending a message with mentalistic semantics (e.g., in KQML~\cite{labrou1997proposal}) in the context of an interaction protocol should recommend the ability of an agent to have mental states, in order to support compliance to the protocol.

An agent may, further, have an \textit{ability to reason about actions and to plan ahead by using specific methods and mechanisms}, e.g., by using a STRIPS-like planner to synthesize plans~\cite{georgeff1987reactive}. Then, a signifier that targets agents with such a planning ability should signify behavior specifications that can contribute to the construction of a suitable planning domain (e.g., specifications of actions with their preconditions and postconditions). Alternatively, a signifier that targets BDI agents that implement the Procedural Reasoning System~\cite{georgeff1987reactive} could instead signify behavior specifications as plans.

An agent also may have an \textit{ability to behave based on its social context}, e.g., to fulfill its role of machine operator by operating  machines in an industrial shopfloor. Recommending role-related abilities within a multi-agent setting (e.g., an organization~\cite{weiss2000multi}) can enable agents to interpret which affordances are exploitable based on a role or help to fulfill a role. For example, a signifier specifying a behavior for operating a robotic arm on a manufacturing shopfloor should recommend the ability of an agent to play the role of a robot operator rather than the role of a warehouse manager. 

An agent also may have an \textit{ability to operate within a given domain}, e.g., by having knowledge of abstractions and processes in industrial manufacturing. For example, a signifier that signifies a behavior specification using a specific semantic model for industrial processes, such as the SAREF4INMA\footnote{SAREF4INMA defines a vocabulary for the industry and manufacturing domain; available online: \url{https://saref.etsi.org/saref4inma/v1.1.2/}} semantic model, should recommend the ability of an agent to interpret the required model. 

An agent, finally, may have an \textit{ability to perform a behavior when an affordance is present}, e.g., to behave based on a behavior specification. Such abilities could originate from the set $Ability\mbox{-}to\mbox{-}B$ and their recommendation by a signifier falls under case (C2). For example, a signifier that signifies a specification for \texttt{close-gripper} (Lst. \ref{lst:sg-basic}) could recommend the ability of an agent to \emph{log in} as operator.

\subsection{Environment Support for Signifier Exposure}
\label{subsec:sem}

Based on our formalization of signifiers, and the discussion of the diverse types of agent abilities that can be considered with our proposed model, we next discuss how the \emph{exposure} of signifiers can be realized and dynamically adjusted in hypermedia environments.

\subsubsection{Exposing Signifiers in the Environment}

Signifiers are not simple informational resources, but rather constructs available in hypermedia environments that enable \textit{situated} agents to discover and interpret affordances. Here, we consider that signifiers can be exposed through \textit{workspaces} that logically partition the environment (similar to A\&A workspaces~\cite{ricci2011environment})\footnote{We formalize workspace to provide environment support for managing interaction cues/metadata through containes (e.g. as in HCI via the ``natural mapping'' principle~\cite{norman2013design}, or in Web-based systems via W3C directories~\cite{Tavakolizadeh:23:WTD} and API ecosystems~\cite{medjaoui2021continuous}). However, we do not expect (or impose) that workspaces is the only means to manage a MAS environment and signifier exposure.}. 

Specifically, a workspace in a hypermedia environment is a container of Web resources and the interactions that are enacted among contained Web resources. A workspace $w$ is formally defined as:
\begin{equation}
    w = (R, I), 
\end{equation}
where $R$ is the set of resources contained in the workspace $w$ (e.g., agents, signifiers, other workspaces etc.) and $I$ is the set of interactions that take place in $w$, enacted among resources of the set $IR \subseteq R$ that offer and exploit affordances.

We further define that $IR=Ag \cup Ar$, where $Ag$ is the set of the agents that are situated in the workspace $w$, and $Ar$ is the set of  the non-autonomous entities, i.e. artifacts~\cite{ricci2011environment}, that are contained in $w$. Then, signifiers are designed to enable the interactions of resources in $IR$, i.e. offered by artifacts in $Ar$ to agents in $Ag$. To examine what the environment affords in the context of agent-to-agent interaction, an affordance offered by an agent $ag \in Ag$ is treated as an affordance of the body\footnote{The body of an agent is an artifact that enables the agent to be situated in a workspace and interact with other resources in the workspace~\cite{ricci2006cartago}.} $body_{ag}$ of $ag$, where $body_{ag} \in Ar$\footnote{Following the definition of affordances (cf. Def.~\ref{eq:aff}), we only consider agent-to-environment (agent-to-artifact) interaction. Agent-to-agent interaction falls under agent-to-environment interaction through agent bodies, and environment-to-environment (artifact-to-artifact) interaction is not considered.}. 

However, the work of designers should not be limited to the construction of signifier tuples as given by Def.~\ref{eq:signifier}. Instead, upon enriching the environment with signifiers, designers should in addition consider how to support the discoverability of signifiers by agents that are expected in the environment \emph{at run time}.\footnote{This represents a significant broadening of the signifier concept with respect to the HCI literature, where signifiers are thought as being able to be modulated \emph{at run time} only in edge cases.}
To advertise signifiers for affordances exposed by an \textit{artifact}, the environment designer can create an \textit{artifact profile},
i.e. structured data describing the artifact through signifiers and general (domain- and application-specific) metadata. Formally, a profile $p_{ar}$ of an artifact $ar$ is defined as the following construct:
\begin{equation}
\label{eq:profile-ar}
    p_{ar} = (Sig_{ar}, s_{ar}), p_{ar} \in R,
\end{equation}
where $Sig_{ar}$ is the set of the signifiers that are exposed in the profile $p_{ar}$, and $s_{ar}$ is metadata capturing part of the situation of the $ar$.

For an artifact $ar$ in a workspace $w$, at least one artifact profile $p_{ar}$ is contained in $w$ to explicitly capture the containment relationship between $ar$ and $w$. 
Then, the discovery of the signifier set $Sig_{ar}$ in $p_{ar}$ is enabled by the containment of $ar$ in $w$.
 
\subsubsection{Dynamically Adjusting Signifier Exposure}

In large-scale environments, additional support for the discoverability of signifiers is needed as the number of artifacts, affordances and, consequently, signifiers grows. At the same time, the interpretability of signifiers may be hindered within open environments where agents interpret signifiers based on diverse abilities. 

To address these issues we propose a \emph{Signifier Exposure Mechanism} (SEM) that exposes a filtered set of signifiers based on the targeted agent's abilities and situation (e.g., the agent's goals), and the situations of artifacts (e.g., based on the valid transitions from the current artifact state).

First, we consider the set $P_{Ag}$ that is the set that contains the profiles of each agent $ag \in Ag$ in the workspace $w$. Then, each profile $p_{ag} \in P_{Ag}$ is metadata describing the abilities and the situation of the agent $ag$. The profile  $p_{ag}$  of the agent $ag$ is then formally defined as the following construct: 
\begin{equation}
    p_{ag} = (A_{ag}, s_{ag}),  p_{ar} \in P_{Ag} \subset R,
\end{equation}
where $A_{ag}$ is the set of the abilities of the agent $ag$, and $s_{ag}$ is metadata capturing part of the situation of the agent $ag$.

Additionally, we consider the set $P_{Ar}$ that is the set that contains the profiles of each artifact $ar \in Ar$ (cf. Def.~\ref{eq:profile-ar}) in the workspace $w$. 

We now present a definition of an SEM (cf.~\cite{lemee2022sign}) whose functionality is described as follows: Given an agent profile $p_{ag} \in P_{Ag}$, and given an artifact profile  $p_{ar} \in P_{Ar}$, the SEM outputs an artifact profile $p_{ar}'$ that exposes only those signifiers of $p_{ar}$ that match a) the abilities of agent $ag$ and b) the agent-environment situation (i.e. the $ag\mbox{-}ar$ situation). 
For the former, we simply consider if the set of the recommended abilities is a subset of the agent's abilities.
For the latter, we use an evaluation function $E$ that evaluates to what degree the situation of the agent and the situation of the artifact conform to the context that is recommended by the signifier. For this, we consider $C_{ar}$ as the set of all the recommended contexts that can be accessed through the profile of artifact $ar$. Then, we use $SC_{Ar}$ to denote the set of all the recommended contexts that can be accessed through the set of artifact profiles $P_{Ar}$, paired with the situations that are captured in their corresponding profiles of $P_{Ar}$. Pairs $(c_{ar}, s_{ar}) \in SC_{Ar}$ are used for associating the situation of an artifact $ar$ only with the contexts that apply to the same artifact $ar$:
    \begin{equation}
        SC_{Ar} =\{ C_{ar} \times \{s_{ar}\}\ \:|\: p_{ar} = (Sig_{ar}, s_{ar}) \land ar \in Ar\}. 
    \end{equation}
We, also, use $S_{Ag}$ to denote the set that includes the situations that can be accessed through the set of agent profiles $P_{Ag}$, formally:
\begin{equation}
    S_{Ag} = \{ s_{ag}\:|\:p_{ag} = (A_{ag}, s_{ag}) \land ag \in Ag \}
\end{equation}
Then, we define the evaluation function $E$ that validates an agent situation of the set $S_{Ag}$ and an artifact situation of the set $CS_{Ar}$ against a context of the set $CS_{Ar}$:
\begin{equation}
\begin{aligned}
    E:  CS_{Ar} \times S_{Ag} &\longrightarrow [0,1].
\end{aligned}
\end{equation}   
Finally, $P_{Ar}'$ is the set of possible produced artifact profiles:
\begin{equation}
\begin{aligned}
   \medmath{P_{Ar}' = }
   & \medmath{\{(Sig_{ar}', s_{ar}) \;|\;  p_{ar} = (Sig_{ar}, s_{ar})}
    \medmath{\land\; Sig_{ar}' \in \mathcal{P}(Sig_{ar}) \;\land\; ar \in Ar \}}
\end{aligned}
\end{equation}
We can now formally define $SEM$ as follows:
\begin{equation}
\label{eq:sem}
\begin{aligned}
   \medmath{ SEM : P_{Ag} \times P_{Ar} \times \{e | e \in [0,1]\}}
   &\medmath{\longrightarrow P_{Ar}'};\\
  \medmath{ ((Sig_{ar}, s_{ar}), (A_{ag}, s_{ag}), t) \longmapsto }
   &\medmath{(\{sig = (sp_b, A, c, salience) \in Sig_{ar} }\\
   &\medmath{|\:  A \subseteq A_{ag} \land E((c, s_{ar}), s_{ag}) > t\}, s_{ar})}
\end{aligned}
\end{equation} 
where $t$ is a threshold value that is given as input to the SEM.

While it may be beneficial, it is not required that exposed signifiers should always relate to affordances that are considered exploitable or relevant. For example, there may be cases where observations about agents and artifacts are insufficient for evaluating relevance, for instance within highly dynamic environments or when an agent prefers not to share such information. Additionally, agents with planning abilities may require an action space that relates to affordances which are not necessary currently exploitable.


\subsection{Customizing Signifiers to Heterogeneous Reasoning Abilities of Agents}\label{subsec:srm}

In this section, we examine how a generic signifier can be extended to reveal information that is relevant to agents with different reasoning and planning abilities. We, specifically, consider the design of signifiers for a) BDI agents that implement the Procedural Reasoning System (PRS), and b) agents capable of planning their actions by using a STRIPS-like planner.

To illustrate our approach with the generic signifier in Lst.~\ref{lst:sg-basic}, which signifies an action specification for closing the gripper of a robotic arm. In the following, we show how this signifier can be extended to customize its content towards accommodating the abilities of the targeted agents.

\subsubsection{Customizing Signifiers for PRS-based Reasoning}

A BDI agent with a PRS-based ability is expected to look for signifiers for satisfying its \textit{intentions}, i.e. the \textit{goals} that the agent has committed to achieve. Although the agent's \textit{plan library} may already contain a plan, i.e. a course of action for accomplishing the agent's goal, a dynamic environment does not always permit for the required actions to be coupled to specific action implementations. More concretely, in case an agent has access to a pick-and-place plan, it may intent to act on a robotic arm (e.g., to close an arm's gripper), but without knowing which robotic arm to use in its current workspace, or any other lower-level interaction details.

To discover fitting signifiers in our approach, an agent may update its profile to describe (part of) its \textit{mental attitudes} -- its current \textit{beliefs} and \textit{goals}, for instance its desire to achieve a specific goal while looking for the appropriate means to satisfy its desire. Considering the desire of an agent to achieve the goal of picking up an item and placing it in a specified location, the agent's profile could be similar to the profile presented in Lst.~\ref{lst:agent}\footnote{We consider that the agent has a desire that is a goal achievement as specified in the AgentSpeak language~\cite{bordini2007programming}
.}. 

\begin{lstlisting}[float= b,floatplacement=H,style=RDFStyle, caption=The resource profile of a BDI agent that implements the PRS and desires to pick and place an item., label=lst:agent, numbers=left, xleftmargin=12pt, mathescape=true]
@base <http://ex.org/wksp/1/arts/2>.
@prefix item-profile: <http://ex.org/wksp/1/arts/3#>
$\lstsetnumber{\ldots}$... $\lstresetnumber\setcounter{lstnumber}{10}$
<> a hmas:AgentProfile ;  hmas:isProfileOf <#agent> . 

<#agent> a hmas:Agent ; 
 hint:hasAbility [ a prs:PRSAbility ] ; 
 hint:hasAbility [ a manu:OperatorAbility ] ;
 prs:hasDesire [ a prs:GoalAchievement,manu:PickAndPlace; 
  prs:hasInputList [ a rdf:List ;
   rdf:first wksp:item ;
   rdf:rest [ a rdf:List ; 
    rdf:first <#location>] ] .
		
item-profile:item a manu:Item ;
 manu:hasLocation item-profile:location .
\end{lstlisting}

\begin{lstlisting}[float= b,style=RDFStyle, caption=A customized signifier for agents that implement a BDI architecture based on the PRS., label=lst:sg-prs, numbers=left, xleftmargin=14pt, mathescape=true ]
$\lstsetnumber{\ldots}$... $\lstresetnumber\setcounter{lstnumber}{8}$
<#sig> a hmas:Signifier ;
 hint:signifies <#close-gripper> ;
 hint:recommendsAbility [ a prs:PRSAbility ] ;
 hint:recommendsAbility [ manu:OperatorAbility ] ;
 hint:recommendsContext <#env-context>, <#prs-context> .

<#prs-context> a hint:Context; sh:targetClass hmas:Agent ; 
 sh:property [ sh:path prs:hasDesire ; 
  sh:minCount 1 ; sh:qualifiedMinCount 1 ; 
  sh:qualifiedValueShape <#desire-shape> ] .

<#desire-shape> a sh:NodeShape ;
 sh:class manu:PickAndPlace; 
 sh:property [ sh:path prs:hasInputList $\lstresetnumber\setcounter{lstnumber}{42}$
   $\lstsetnumber{\ldots}$ ... ] .$\lstresetnumber\setcounter{lstnumber}{43}$

<#item-shape>  a sh:NodeShape ;
 sh:class manu:Item ;
 sh:property [ sh:path manu:hasLocation ; $\lstresetnumber\setcounter{lstnumber}{58}$
  $\lstsetnumber{\ldots}$ ... ] .  $\lstresetnumber\setcounter{lstnumber}{59}$
 
<#location-shape> a sh:NodeShape ;
 sh:class manu:Location ; 
 sh:property [ sh:path manu:inRangeOf ;
  sh:minCount 1 ;
  sh:hasValue ex:leubot ] .
\end{lstlisting}
Taking into account the abilities of PRS-based agents, designers can now construct signifiers recommending contexts that relate to the expected situations of the targeted agents, i.e. situations in terms of \textit{beliefs}, \textit{desires}, and \textit{intentions}. Lst.~\ref{lst:sg-prs} is an example of a signifier for the affordance \texttt{gripper-closable} of a robotic arm, which \emph{extends} the generic signifier for the same affordance (Lst.~\ref{lst:sg-basic}) to better cater to such agents. In this case, the recommended context is a set of constraints that should be validated by the agent-environment situation in order for the signifier to be considered relevant for the agent. The constraints are expressed in the Shapes Constraint Language (SHACL)\footnote{The SHACL specification is available online: \url{https://www.w3.org/TR/shacl/}} and concern the following aspects of the agent-artifact situation: 1) the agent has the desire to pick an item and place it in a target location (l.~20-43);
2) the item is in range of the robotic arm that offers the affordance \texttt{gripper-closable} (l.~45-59, 61-65); 
3) the target location is in the range of the robotic arm that offers the affordance \texttt{gripper-closable} (l.~45-59, 61-65).

Additionally, the signifier is extended to recommend a \emph{PRS-based ability}. Therefore, if an agent with such an ability looks up signifiers at run time, and an SEM is available (see Sect.~\ref{subsec:sem}), the SEM will expose the \emph{customized} signifier of Lst.~\ref{lst:sg-prs} rather than the generic signifier of Lst.~\ref{lst:sg-basic}. Finally, if the SEM has a SHACL validation feature, exposure will be adjusted based on whether the agent's situation (Lst.~\ref{lst:agent}) conforms to the SHACL context shape\footnote{An SEM with a SHACL processor has already been presented in \cite{taghzouti2022step}. There, agents specify SHACL shapes that impose constraints on signifiers, and they provide such shapes as input to the SEM upon looking for conforming signifiers. On the other hand, here, we consider that environment designers create the signifiers and specify constraints for identifying at run time conforming agent-environment situations.}. In this case, the customized signifiers will be exposed to agents with a PRS-based ability only if the agent-environment situation conforms to the recommended constraints. 

\begin{lstlisting}[float=b,style=RDFStyle, caption=A customized signifiers for agents with a STRIPS planning ability., label=lst:sg-strips, numbers=left, xleftmargin=14pt, mathescape=true ]
@prefix pddl: 
<http://www.cs.yale.edu/homes/dvm/daml/pddlonto.daml#>.
$\lstsetnumber{\ldots}$... $\lstresetnumber\setcounter{lstnumber}{8}$
<#sig> a hmas:Signifier ;
 hint:signifies <#close-gripper> ;
 hint:recommendsAbility [ 
  a strips:StripsPlanningAbility ] .

<#close-gripper> a hint:ActionSpecification;
$\lstsetnumber{\ldots}$... $\lstresetnumber\setcounter{lstnumber}{20}$
 a a pddl:Action ;
 pddl:action-label "closeGripper";
 pddl:parameters [ a pddl:Param_seq ;
  rdf:_1 <#param1> ];
 pddl:precondition [$\lstresetnumber\setcounter{lstnumber}{30}$
  $\lstsetnumber{\ldots}$... ] ; $\lstresetnumber\setcounter{lstnumber}{31}$
 pddl:effect [$\lstresetnumber\setcounter{lstnumber}{40}$
  $\lstsetnumber{\ldots}\lstresetnumber\setcounter{lstnumber}{41}$... ] .
  
<#param1> a pddl:Param ; 
 pddl:name "?gv" ;
 drs:type manu:GripperValue ;
 :hasSchema <#gripper-schema> .
\end{lstlisting}

\subsubsection{Customizing Signifiers for STRIPS-based Reasoning}

Instead of profiting from signifiers that are designed with agents' beliefs, desires, and intentions in mind, agents that are capable of automated planning would require signifiers that signify action specifications suitable to enrich a planning domain.  
For instance, Lst.~\ref{lst:sg-strips} presents a customized signifier for the affordance \texttt{gripper-closable} of a robotic arm, which extends the generic signifier of Lst.
~\ref{lst:sg-basic} to signify an action specification that is relevant to agents with a STRIPS-based planning ability. The action specification specifies the type of the action that can be performed, and, additionally, the \textit{preconditions} and the \textit{effects} of executing the action. For this example, we reused the PDDL ontology presented in~\cite{mcdermott2002representing} to specify actions that can be used to enrich a PDDL domain and, thus, become suitable input to a PDDL automated planner. The signifier is also extended to recommend a \textit{STRIPS planning ability}, so that an SEM can adjust the signifier exposure based on the abilities of the requesting agent. 

\section{IMPLEMENTATION AND DEPLOYMENT}

To ground our approach in a concrete use case, we present a prototypical Hypermedia MAS and a demonstrator scenario.

\subsection{Prototype Deployment}
\label{subsec:sysarch}

In our implementation~\footnote{The implementation is available online: \url{https://github.com/Interactions-HSG/} \url{yggdrasil/tree/feature/sem}}, we use the Yggdrasil platform for Hypermedia MAS~\cite{ciortea2018engineering} as a repository for our hypermedia environment. We extended the Yggdrasil API to enable the publication of agent and artifact profiles as presented in Sect.~\ref{subsec:signifiers}. The API hence permits the publication and management of resource profiles based on a) the Hypermedia MAS Core Ontology (HMAS-core) that is used to describe core aspects of Hypermedia MAS (e.g., workspaces, resource profiles etc.), and b) the Hypermedia MAS Interaction ontology which extends HMAS-core based on our model of signifiers. 

We further extended Yggdrasil with an SEM for dynamically adjusting the signifier exposure based on the formal model presented in Sect.~\ref{subsec:sem}. Specifically, agent and artifact profiles can be published so that they are accessible through workspaces of the hypermedia environment. As a result, when an agent discovers an artifact profile in the hypermedia environment, the SEM is responsible for evaluating which signifiers should be exposed to the agent: The SEM identifies which agent is currently looking for signifiers, and attempts to retrieve the agent's profile. If no agent profile is available, the artifact profile is provided to the agent without undergoing any signifier adjustment. On the other hand, if the SEM acquires access to an agent profile, it proceeds to evaluate the complementarity between the agent and the artifact as given by Def.~\ref{eq:sem}. For example, if the agent exhibits specific abilities, then the SEM will construct a variation of the artifact profile that exposes only these signifiers that contain recommendations for the specified agent abilities.

For our prototype, we used Jason~\cite{bordini2007programming} agents which were implemented and deployed in a JaCaMo\footnote{The JaCaMo documentation is available online: \url{https://jacamo.sourceforge.net/}} application. The signifiers used were designed for revealing information about affordances of a PhantomX AX12 Reactor Robot Arm\footnote{\url{https://www.trossenrobotics.com/p/phantomx-ax-12-reactor-robot-arm.aspx}}.  

\subsection{Deployment Scenario}

To validate our prototype implementation, we considered the case of a Hypermedia MAS where affordance discovery and exploitation take place in a workspace with industrial devices based on the following scenario: A manufacturing workspace contains a robotic arm that offers affordances to agents that are situated in the workspace, such as affordances to move the gripper and the base of the robotic arm. The robotic arm is modelled as an \textit{artifact} and its presence is indicated in the hypermedia environment through its \textit{artifact profile} that is contained in the workspace. Signifiers that reveal information about the affordances of the robotic arm are exposed through its profile, and have been designed by taking into consideration different abilities of agents that are expected to be situated in the manufacturing workspace. Specifically, we have considered three types of signifiers with regard to agents' abilities:
\begin{itemize}
    \item Signifiers that recommend the abilities of BDI agents that implement the Procedural Reasoning System (PRS).
    \item Signifiers that recommend the abilities of agents that can perform automated planning using a STRIPS-like planner. 
    \item Generic signifiers that are not customized to agents with any specific abilities of reasoning about action and acting.
\end{itemize}

A BDI agent that implements the PRS joins the workspace with the objective to pick and place an item. The agent's presence in the workspace is indicated in the hypermedia environment through its profile, i.e. metadata describing that a) the body of the agent is contained in the workspace, b) the agent desires to \texttt{pick-and-place}, and c) the agent has a PRS ability. The agent has already access to a relevant pick-and-place plan through its plan library, however, it is unaware of any lower-level implementation details that are required for effectively executing its plan. For this, the agent looks for signifiers to acquire information about the hypermedia controls that can be used to exploit relevant affordances. Upon looking for information about the affordances of the robotic arm that is contained in the manufacturing workspace, the agent receives only those signifiers that are relevant to its abilities from the SEM. 

Another agent joins the workspace with the same objective of picking and placing an item. Although the agent does not currently have a relevant \texttt{pick-and-place} plan, it has access to a STRIPS-like planner for performing automated planning -- an ability that is indicated in the agent's profile. The agent decides to look for signifiers that may help it to synthesize a pick-and-place plan. Upon focusing on the robotic arm, the agent receives the profile of the robotic arm that exposes only signifiers that are relevant to the agent due to its STRIPS planning ability. Consequently, the agent acquires access to action specifications that would be suitable for enriching a planning domain as they specify the types of actions that can be performed upon the robotic arm, along with the preconditions and effects of executing such actions. 

\section{CONCLUSIONS}

In this paper, we introduced signifiers as a first-class abstraction in Hypermedia MAS.
This was accomplished through formalizations for the design of signifiers that enable the hypermedia-driven exploitation of affordances on the Web, and that capture the agent-environment complementarity required to exploit affordances.
These signifiers are coupled with a mechanism that describes the dynamic adjustment of signifier exposure based on the agent-environment situation.
We provided examples and demonstrated how signifiers decouple behaviors from their implementations through hypermedia, and can be customized with respect to different reasoning abilities of agents, towards enabling the effective and efficient affordance discovery and exploitation in affordance-rich and open Web-based MAS. Based on the presented work, we aim at supporting and evaluating the effectiveness and efficiency of more complex behaviors through the design and exposure of signifiers based on the HATEOAS principle and the behavior-ability impredicativity of Affordance Theory.

\begin{acks}

This research has received funding from the Swiss National Science Foundation under grant No. 189474 (\textit{HyperAgents}) and from the European Union’s Horizon 2020 research and innovation program under grant No. 957218 (\textit{IntellIoT}). We thank Yousouf Taghzouti for his valuable input on the topic of semantic content negotiation.

\end{acks}



\bibliographystyle{ACM-Reference-Format} 
\bibliography{main}


\end{document}